\documentclass[a4paper]{llncs}
\usepackage{graphicx}
\usepackage{latexsym}

\usepackage{times}
\usepackage{soul}
\usepackage{url}
\usepackage[hidelinks]{hyperref}
\usepackage[utf8]{inputenc}
\usepackage[small]{caption}
\usepackage{graphicx}
\usepackage{amssymb,amsfonts,amsmath}
\usepackage{algorithm}
\usepackage{algorithmic}
\newcommand{\KILL}[1]{}
\newcommand{\HIDE}[1]{}
\newcommand{\task}[1]{{\small {\tt #1}}}

\newcommand{\nat}{\mathbb{N}}

\newcommand{\runningtask}{\task{b94a9452}}

\begin{document}

\title{Tackling the Abstraction and Reasoning Corpus (ARC)\\ with Object-centric Models and the MDL Principle}

\author{Sébastien Ferré}

\institute{Univ Rennes, CNRS, Inria, IRISA\\
  Campus de Beaulieu, 35042 Rennes, France\\
  Email: \email{ferre@irisa.fr}}

\maketitle

\begin{abstract}
  The Abstraction and Reasoning Corpus (ARC) is a challenging
  benchmark, introduced to foster AI research towards human-level
  intelligence. It is a collection of unique tasks about generating
  colored grids, specified by a few examples only. In contrast to the
  transformation-based programs of existing work, we introduce
  object-centric models that are in line with the natural programs
  produced by humans. Our models can not only perform predictions, but
  also provide joint descriptions for input/output pairs. The Minimum
  Description Length (MDL) principle is used to efficiently search the
  large model space. A diverse range of tasks are solved, and the
  learned models are similar to the natural programs. We demonstrate
  the generality of our approach by applying it to a different domain.
\end{abstract}

\section{Introduction} 

Artificial Intelligence (AI) has made impressive progress in the past
decade at specific tasks, sometimes achieving super-human performance:
e.g., image recognition~\cite{KriSutHin2012nips}, board
games~\cite{Silver2016alphago}, natural language
processing~\cite{BERT2019}. However, AI still misses the generality
and flexibility of human intelligence to adapt to novel tasks with
little training. To foster AI research beyond narrow
generalization~\cite{Goertzel2014},
F. Chollet~\cite{Chollet2019,Chollet2020} introduced a measure of
intelligence that values {\em skill-acquisition efficiency} over {\em
  skill performance}, i.e. the amount of prior knowledge and
experience that an agent needs to reach a reasonably good level at a
range of tasks (e.g., board games) matters more than its absolute
performance at any specific task (e.g., chess).
Chollet also introduced the Abstraction and Reasoning
Corpus (ARC) benchmark in the form of a psychometric test to measure
and compare the intelligence of humans and machines alike. ARC is a
collection of tasks that consist in learning how to transform an input colored
grid into an output colored grid, given only a few examples.
It is a very challenging benchmark. While humans can solve more than
80\% of the tasks~\cite{Johnson2021}, the winner of a Kaggle
contest\footnote{\url{https://www.kaggle.com/c/abstraction-and-reasoning-challenge}}
could only solve 20\% of the tasks (with a lot of hard-coded
primitives and brute-force search), and the winner of the more recent
ARCathon'22
contest\footnote{\url{https://lab42.global/past-challenges/arcathon-2022/}}
could only solve 6\% of the tasks.

The existing published
approaches~\cite{Fischer2020,Alford2021,Xu2022,Ainooson2023}, and also
the Kaggle winner, tackle the ARC challenge as a {\em program
  synthesis} problem, where a program is a composition of primitive
transformations, and learning is done by searching the large program
space.
In contrast, psychological studies have shown that the {\em natural
  programs} produced by humans to solve ARC tasks are object-centric,
and more declarative than procedural~\cite{Johnson2021,Acquaviva2022}.
When asked to verbalize instructions on how to solve a task,
participants typically first describe what to expect in the input
grid, and then how to generate the output grid based on the elements
found in the input grid.

We make two contributions w.r.t. existing work:
\begin{enumerate}
\item {\em object-centric models} that enable to both parse and
  generate grids in terms of object patterns and computations on those
  objects;
\item an efficient search of object-centric models based on the {\em
    Minimum Description Length (MDL) principle}~\cite{Rissanen1978}.
\end{enumerate}
A model for an ARC task combines two {\em grid models}, one for the
input grid, and another for the output grid. This closely matches the
structure of natural programs. Compared to the transformation-based
programs that can only predict an output grid from an input grid, our
models can also provide a joint description for a pair of grids. They
can also create new pairs of grids, although this is not evaluated in
this paper. They could also in principle be adapted to tasks based on
a single grid or on sequences of grids. All of this is possible
because grid models can be used both for parsing a grid and for
generating a grid.

The MDL principle comes from information theory, and says that {\em
  ``the model that best describes the data is the model that compress
  them the more''}~\cite{Rissanen1978,Grunwald2019}.  It has for
instance been applied to pattern mining~\cite{KRIMP2011,Vouw2020}.
The MDL principle is used at two levels: (a) to choose the best parses
of a grid according to a grid model, and (b) to efficiently search the
large model space by incrementally building more and more accurate
models.
MDL at level (a) is essential because the segmentation of a grid into
objects is task-dependent and has to be learned along with the
definition of the output objects as a function of the input objects.
The two contributions support each other because existing search
strategies could not handle the large number of elementary components
of our grid models, and because the transformation-based programs are
not suitable to the incremental evaluation required by MDL-based
search.

We report promising results based on grid models that are still far
from covering all knowledge priors assumed by ARC. Correct models are
found for 96/400 varied training tasks with a 60s time budget. Many of
those are similar to the natural programs produced by
humans. Moreover, we demonstrate the generality of our approach by
applying it to the automatic filling of spreadsheet
columns~\cite{Gulwani2011}, where inputs and outputs are rows of
strings instead of grids.

The paper is organized as follows. Section~\ref{arc} presents the ARC
benchmark, and a running example task. Section~\ref{related} discusses
related work. Section~\ref{models} defines our object-centric models,
and Section~\ref{learning} explains how to learn them with the MDL
principle. Section~\ref{eval} reports on experimental results,
comparing with existing approaches.

\section{Abstraction and Reasoning Corpus (ARC)} 
\label{arc}

\begin{figure}[t]
  \centering
  \includegraphics[width=\columnwidth]{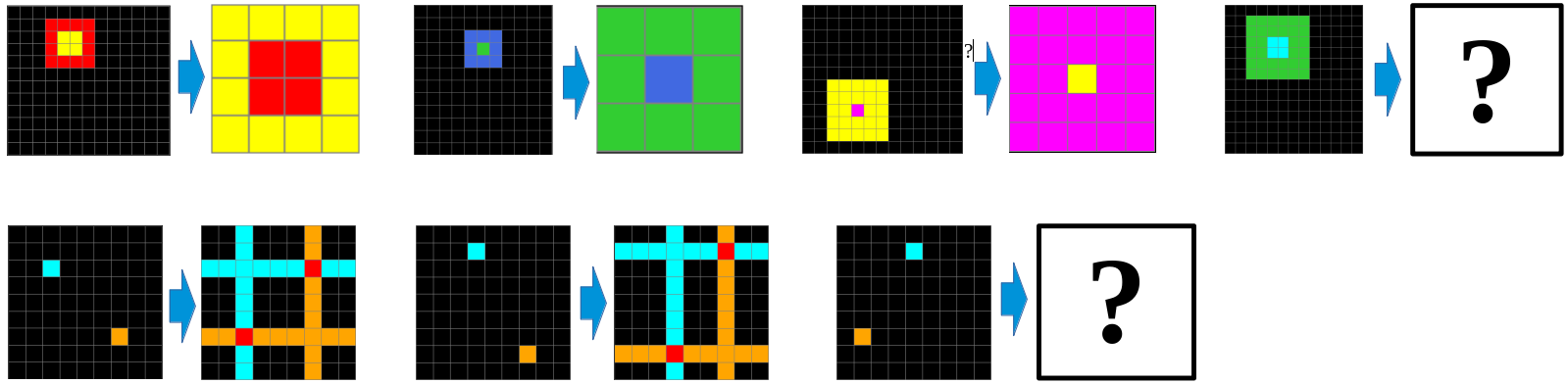}
  \caption{Training tasks \task{b94a9452} (top) and \task{23581191}
    (bottom), with 2-3 demonstration examples (left) and the input grid of
    a test case (right).}
  \label{fig:task}
\end{figure}

ARC is a collection of tasks\footnote{Data and testing interface at
  \url{https://github.com/fchollet/ARC}}, where each task is made of
training examples (3.3 on average) and test examples (1 in
general). Each example is made of an input grid and an output
grid. Each grid is a 2D array (with size up to 30x30) filled with
integers coding for colors (10 distinct colors). For a given task, the
size of grids can vary from one example to another, and between the
input and the output.
Each task is a machine learning problem, whose goal is to learn a
model that can generate the output grid from the input grid, and so
from a few training examples only. Prediction is successful only if
the predicted output grid is {\em strictly equal} to the expected grid
for {\em all} test examples, there is no partial success. However,
three trials are allowed for each test example to compensate for
potential ambiguities in the training examples.
Figure~\ref{fig:task} shows two ARC tasks (with the expected test
output grid missing). The first is used as a running example in this
paper.

We now more formally define grids, examples, and tasks.

\begin{definition}[grid]
  A {\em grid} $g \in C^{h \times w}$ is a matrix with values taken
  from a set of colors~$C$, with $h>0$ rows ({\em height}), and $w>0$
  columns ({\em width}).
  A {\em grid cell} is identified by {\em coordinates}~$(i,j)$, where
  $i$ selects a row, and $j$ selects a column. The color at
  coordinates~$(i,j)$ is denoted by either~$g_{ij}$ or~$g[i,j]$.
  Coordinates range from~$(0,0)$ to~$(h-1,w-1)$.
\end{definition}

ARC grids use 10 colors, and have height/width up to 30.

\begin{definition}[example] 
  An {\em example} is a pair of grids \mbox{$e = (g^i,g^o)$}, where $g^i$ is called
  the {\em input grid}, and $g^o$ is called the {\em output grid}.
\end{definition}

As illustrated by Figure~\ref{fig:task}, the output grid needs not
have the same size as the input grid, it can be smaller or bigger.

\begin{definition}[task]
  A {\em task} is a pair~$T = (E,F)$, where $E$ is the set of {\em
    training examples}, and $F$ is the set of {\em test examples}.
\end{definition}

ARC tasks have 3.3 training examples on average, and 1 or 2 test
examples (most often 1).
As illustrated by Figure~\ref{fig:task}, the different input grids of
a task need not have the same size, nor use the same colors. The same
applies to test grids.

ARC is composed of 1000 tasks in total: 400 ``training
tasks''\footnote{The term ``training tasks'' may be misleading as
  their purpose is to train AI developers, not AI systems. Humans solve ARC tasks without training.},
400 evaluation tasks, and 200 secret tasks for independent
evaluation.
Figure~\ref{fig:task} shows two of the 400 training tasks. Developers
should only look at the training tasks, not at the evaluation
tasks. The latter should only be used to evaluate the broad
generalization capability of the developed systems.

\section{Related Work}
\label{related}

The ARC benchmark is recent and not many approaches have been
published so far. All those we know define a DSL (Domain-Specific
Language) of programs that transform an input grid into an output
grid, and search for a program that is correct on the training
examples~\cite{Fischer2020,Alford2021,Xu2022,Ainooson2023}. The
differences mostly lie in the primitive transformations (prior
knowledge) and in the search strategy. It is tempting to define more
and more primitives like the Kaggle winner did, hence more prior
knowledge, but this means a less intelligent system according to
Chollet's measure. To guide the search in the huge program space,
those approaches use either grammatical evolution~\cite{Fischer2020},
neural networks~\cite{Alford2021}, search tree pruning with hashing
and Tabu list~\cite{Xu2022}, or stochastic search trained on solved
tasks~\cite{Ainooson2023}.
A difficulty is that the output grids are generally only used to score
a candidate program so that the search is kind of
blind. Alford~\cite{Alford2021} improves this with a neural-guided
bi-directional search that grows the program in both directions, from
input and output. Xu~\cite{Xu2022} compares the in-progress generated
grid to the expected grid but this limits the approach to the tasks
whose output grids have the same size and same objects as the input
grids. DSL-based approaches have a scaling issue because the search
space increases exponentially with the number of
primitives. Ainooson~\cite{Ainooson2023} alleviates this difficulty by
defining high-level primitives that embody specialized search
strategies.
We compare and discuss their performance in the evaluation section.


Johnson {\em et al.}~\cite{Johnson2021} report on a psychological
study of ARC. It reveals that humans use object-centric mental
representations to solve ARC tasks. This is in contrast with existing
solutions that are based on grid transformations. Interestingly, the
tasks that are found the most difficult by humans are those based on
logics (e.g., an exclusive-or between grids) and symmetries (e.g.,
rotation), precisely those most easily solved by DSL-based approaches.
The study exhibits two challenges: (1) the need for a large set of
primitives, especially about geometry; (2) the difficulty to identify
objects, which can be only visible in part due to overlap or
occlusion.
A valuable resource is LARC, for Language-annotated
ARC~\cite{Acquaviva2022}, collected by crowd-sourcing. It provides for
most training tasks one or several {\em natural programs} that confirm
the object-centric and declarative nature of human representations.
A natural program is short textual descriptions produced by a
participant that could be used by another participant to generate test
output grids (without access to the training examples).


Beyond the ARC benchmark, a number of work has been done in the domain
of {\em program synthesis}, which is also known as program induction
or programming by examples (PbE)~\cite{Lieberman2001pbe}.
An early approach is Inductive Logic
Programming (ILP)~\cite{MugRae1994}, where target predicates are
learned from symbolic representations.
PbE is used in the FlashFill feature of Microsoft Excel 2013 to learn
complex string processing formulas from a few
examples~\cite{FlashFill2013}. Dreamcoder~\cite{Ellis2021dreamcoder}
alternates a {\em wake} phase that uses a neurally guided search to
solve tasks, and a {\em sleep} phase that extends a library of
abstractions to compress programs found during wake.
Bayesian program learning was shown to outperform deep learning at parsing and generating handwritten world's alphabets~\cite{LakSalTen2015}.


\section{Object-centric Models for ARC Grids}
\label{models}

We introduce {\em object-centric models} as a mix of patterns and
functions, in contrast to DSL-based programs that are only made of
functions. We examplify them with {\em grid models} that describe
ARC grids in term of objects having different shapes, colors, sizes, and
positions.
Such grid models are used to {\em parse} a grid, i.e. to understand
its contents according to the model, and also to {\em generate} a
grid, using the model as a template.
A {\em task model} comprises two grid models that enable to predict an
output grid, to describe a pair of grids, or to create a new pair
of grids for the given task.

\subsection{Mixing Patterns and Functions}

The purpose of a grid model is to distinguish between invariant and
variant elements across the grids of a task. In task \runningtask{}
(Figure~\ref{fig:task} top), all input grids contain a square but the
size, color, and position vary. This can be expressed by a {\em
  pattern} {\bf Square}(size$:$?, color$:$?, pos$:$?), where {\bf
  Square} is called a {\em constructor} (here with three arguments),
and the question marks are called {\em unknowns} (similar to Prolog
variables). There is also a constructor for positions as 2D vectors
{\bf Vec}(i$:$?, j$:$?), and primitive values for sizes (e.g., 3) and
colors (e.g., blue).
Patterns can be nested, like in {\bf Square}(3,?,{\bf Vec}(?,2)),
which means {\em ``a square whose size is 3, and whose top left corner
  is on column 2''}, in order to have models as specific as
necessary. Fully grounded patterns (without unknowns) are called {\em
  descriptions}: e.g., {\bf Square}(3,blue,{\bf Vec}(2,4)).

However, with patterns only, there is no way to make the output grid
depend on the input grid, which is key to solving ARC tasks.
We therefore add two ingredients to grid models (typically to output
models): {\em references} to the components of a grid description
(typically the input one), and {\em function} applications to allow
some output components to be the result of a computation. For example,
in task \runningtask, the model for the small square in the output
grids could be {\bf Square}(!small.size, !large.color, !small.pos -
!large.pos), where for instance $!small.size$ is a reference to the
size of the small square in the input grid, and '-' is the
substraction function. This model says that {\em ``the small output
  square has the same size as the small input square, the same color
  as the large input square, and its position is the difference
  between the positions of the two input squares.''}


\begin{table}[t]
  \centering
  \caption{Pattern constructors by type}
  \begin{tabular}{|l|l|}
    \hline
    type & constructors \\
    \hline
    \hline
    {\it Grid} & {\bf Layers}(size$:$ {\it Vector},\ color$:$ {\it Color},\ layers$:$ {\it Layer}[]) \\
         & {\bf Tiling}(grid$:$ {\it Grid},\ size$:$ {\it Vector}) \\
    \hline
    {\it Layer} & {\bf Layer}(pos$:$ {\it Vector},\ object$:$ {\it Object}) \\
    \hline
    {\it Object} & {\bf Colored}(shape$:$ {\it Shape}, color$:$ {\it Color}) \\
    \hline
    {\it Shape} & {\bf Point} \\
         & {\bf Rectangle}(size$:$ {\it Vector},\ mask$:$ {\it Mask}) \\
    \hline
    {\it Mask} & {\bf Bitmap}(bitmap: {\it Bitmap}) \\
         & {\bf Full}, {\bf Border}, {\bf EvenCheckboard}, {\bf OddCheckboard}, ... \\
    \hline
    {\it Vector} & {\bf Vec}(i$:$ {\it Int},\ j$:$ {\it Int}) \\
    \hline
\end{tabular}
\label{fig:model:constructors}
\end{table}

\begin{table}[t]
  \centering
  \caption{Functions by domain}
  \begin{tabular}{|p{0.9\columnwidth}|}
    \hline
    {\bf Arithmetics}: addition and substraction; product and division by a small constant ($2..3$); minimum, maximum and average of two integers; span between two positions ($|x-y|+1$); vectorized versions of the previous functions (e.g., $(i_1,j_1)+(i_2,j_2)=(i_1+i_2,j_1+j_2)$); projection of a vector on an axis. \\
    \hline
    {\bf Geometry}: size and area of an object/shape; extremal and median positions of an object along each axis (e.g., top and bottom, middle); stripping a grid from some background color; cropping a grid at some frame; translation vector of an object against another; scaling an object by a constant factor or relative to a size vector; extension of an object/shape to some size, in agreement to a periodic pattern (e.g., checkerboard); tiling an object/shape along the two axes; applying symmetries to objects/shapes (combining rotations and reflections). \\
    \hline
    {\bf Other functions}: recoloring an object; swapping two colors; color counts and majority color; bitwise operations on masks. \\
    \hline
  \end{tabular}
  \label{fig:model:functions}
\end{table}

Tables~\ref{fig:model:constructors} and \ref{fig:model:functions}
respectively list the pattern constructors and the functions of the
grid models that we have used in our experiments. Each
constructor/function has a result type, and typed arguments. The
argument types constrain which values/constructors/functions can be
used in arguments. The names of constructor arguments are used to
reference the components of a grid model or grid description. Grid,
object and shape constructors have an implicit argument {\em grid} for
their representation as a raw grid. Point shapes have an implicit
argument {\em size}, equal to {\bf Vec}(1,1).
Our grid models describe a grid as either a stack of layers on top of
a background having some size and color, or as the tiling of a grid up
to covering a grid of given size. A layer is an object at some
position. An object is so far limited to a one-color shape, where a
shape is either a point or some mask-specified shape fitting into a
rectangle of some size. A mask is either specified by a bitmap or by
one of a few common shapes such as a full rectangle or a rectangular
border. Positions and sizes are 2D integer vectors.
Four primitive types are used: integers, colors, bitmaps (i.e.,
Boolean matrices), and grids (i.e., color matrices).
The available functions essentially cover arithmetic operations on
integers and on vectors, where vectors represent positions, sizes, and
moves; and geometric notions such as measures (e.g., area),
translations, symmetries, scaling, and periodic patterns (e.g.,
tiling).
Unknowns are here limited to primitive types and vectors. References
and functions are so far only used in output grid models. They could
be used in the input models to express constraints, e.g. to state that
different objects have the same color.

\begin{figure}[t]
\setlength\tabcolsep{2pt}
{\small
\begin{tabular}{ll}
  $M^i =$ & {\bf Layers}(?, black, [\\
          & \hspace{1em} {\bf Layer}(?, {\bf Colored}({\bf Rectangle}(?, {\bf Full}), ?)),\\
          & \hspace{1em} {\bf Layer}(?, {\bf Colored}({\bf Rectangle}(?, {\bf Full}), ?)) ]) \\
  $M^o =$ & {\bf Layers}(!lay[1].object.shape.size, !lay[0].object.color, [\\
          & \hspace{1em} {\bf Layer}(!lay[0].pos - !lay[1].pos,\\
          & \hspace{2em} {\it coloring}(!lay[0].object, !lay[1].object.color)) ]) \\
\end{tabular}
}
\caption{A correct model for task \runningtask.}
\label{fig:model}
\end{figure}

A {\em task model}~$M = (M^i,M^o)$ is made of an input grid
model~$M^i$ and an output grid model~$M^o$. Figure~\ref{fig:model}
shows a correct model for task~\runningtask, which in words says: {\em
  ``There are two stacked full rectangles on a black background in the
  input grid. The size of the output grid is the same as the bottom
  object (lay[1].object), and its background color is the color of the
  top object (lay[0].object). The output grid has a copy of the top
  object, recolored in the color of the bottom object, and whose
  position is the difference between the top object position and the
  bottom object position.''}

\subsection{Parsing and Generating Grids with a Grid Model}

We introduce two operations that must be defined for any grid
model~$M$: the {\em parsing} of a grid~$g$ into a description~$\pi$
and the {\em generation} of a grid description~$\pi$, and thus of a
grid~$g$. These operations are analogous to the parsing and generation
of sentences from a grammar, where syntactic trees correspond to our
descriptions~$\pi$.

In both operations, the references present in the model~$M$ are first
resolved using a description as the evaluation context, called {\em
  environment} and written~$\varepsilon$. Concretely, each reference
is a path in~$\varepsilon$ and is replaced by the sub-description at
the end of this path. The functions applying to these references are
then evaluated. The result is a reduced model~$M'$ consisting only of
patterns and values.

\begin{figure}[t]
   \centering
   \includegraphics[width=0.9\columnwidth]{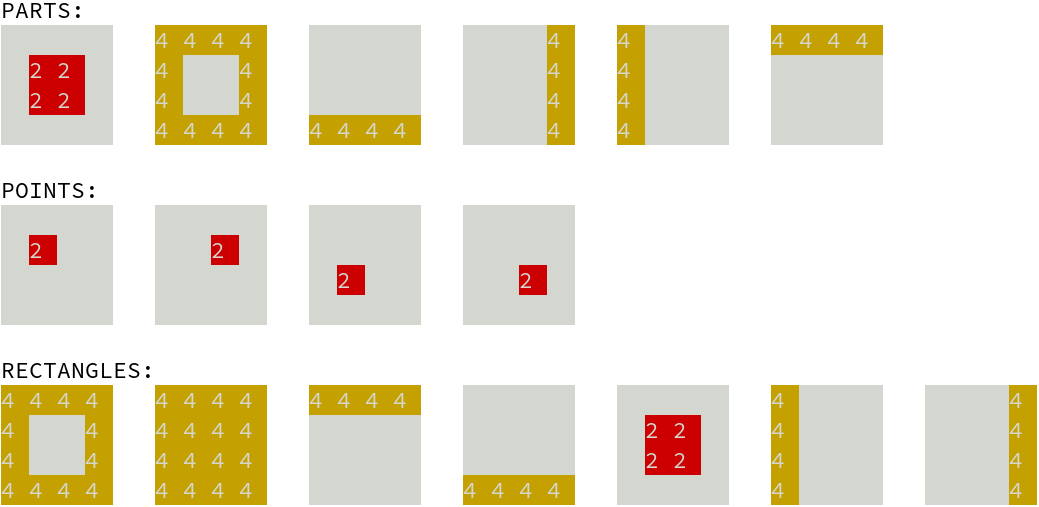}
   \caption{Parts, points and rectangles found in the first output grid}
   \label{fig:parts}
 \end{figure}

{\it Parsing.} The parsing of a grid~$g$ consists in replacing
the unknowns of the reduced model~$M'$ by descriptions corresponding
to the content of the grid. It is not necessary that the whole content
of the grid be described, which allows for partial models. A grid is
analyzed from the top layer to the bottom layer to take into account
overlapping objects. The analysis of an object is contextual, it
depends on what remains to be covered in the grid after the analysis
of the upper layers. For efficiency reasons, each grid is
pre-processed to extract a collection of single-colored parts and the
objects are parsed as unions of these
parts (see Figure~\ref{fig:parts}).
As the analysis of the grids can become combinatorial, we bound the
number of descriptions produced by the parsing and we order them
according to the description length measures defined in
Section~\ref{learning}.
As an example, the parsing of the first input grid of the running task
with the model~$M^i$ of Figure~\ref{fig:model} returns the following
description:
$\pi^i =$ {\small {\bf Layers}({\bf Vec}(12,13), black, [{\bf Layer}({\bf Vec}(2,4), {\bf Colored}({\bf Rectangle}({\bf Vec}(2, 2), {\bf Full}), yellow), {\bf Layer}({\bf Vec}(1,3), {\bf Colored}({\bf Rectangle}({\bf Vec}(4,4), {\bf Full}), red) ])}.

{\it Generation.} The generation of a grid consists in replacing the
remaining unknowns in the reduced model~$M'$ by random descriptions of
the right type, in order to obtain a grid description, which can then
be converted into a concrete grid.
For example, the output model~$M^o$ of Figure~\ref{fig:model} applied
with, as environment~$\varepsilon$, the above description~$\pi^i$ of the
first input grid generates the following description $\pi^o =$
{\small {\bf Layers}({\bf Vec}(4,4), yellow, [{\bf Layer}({\bf
    Vec}(1,1), {\bf Colored}({\bf Rectangle}({\bf Vec}(2,2), {\bf
    Full}), red)])}. This description conforms to the expected output
grid.

An important point is that these two operations are {\em
  multi-valued}, i.e. may return multiple descriptions. Indeed, there
are often several ways of parsing a grid according to a model, for
example if the the model mentions a single object while the grid
contains several ones. There are also several grids that can be
generated by a model when it contains unknowns.

\subsection{Predict, Describe, and Create with Task Models}

We demonstrate the versatility of task models by showing that they can
be used in three different modes: to {\em predict} the output grid
from the input grid, to {\em describe} a pair of grids jointly, or to
{\em create} a new pair of grids for the task.
We use below the notation $\rho, \pi \in {\it parse}(M,\varepsilon,g)$
to say that $\pi$ is the $\rho$-th parsing of the grid~$g$ according
to the model~$M$ and with the environment~$\varepsilon$; and the
notation $\rho,\pi,g \in {\it generate}(M,\varepsilon)$ to say that
$\pi$ is the $\rho$-th description generated by the model~$M$ with the
environment~$\varepsilon$, and that $g$ is the concrete grid described
by~$\pi$. The rank~$\rho$ is motivated by the fact that parsing and
generation are multi-valued.

The {\em predict} mode is used after a model has been learned, in the
evaluation phase with test cases, by predicting an output grid for the
given input grid. It consists in first parsing the input grid with
the input model and the ${\it nil}$ environment in order to get an
input description~$\pi^i$, and then to generate the output grid by
using the ouput model and the input grid description as the
environment.
\begin{align*}
  {\it predict}(M,g^i) = \{(\rho^i,\rho^o,g^o) \mid\ & \rho^i,\pi^i \in {\it parse}(M^i,{\it nil},g^i),\\
                                    & \rho^o,\pi^o,g^o \in {\it generate}(M^o,\pi^i) \}
\end{align*}

The {\em describe} mode is used in the learning phase of the model
(see Section~\ref{learning}). It allows to obtain a joint description
of a pair of grids. It consists in the parsing of the input grid and
the output grid. Let us note that the parsing of the output grid
depends on the result of the parsing of the input grid, hence the term
"joint description".
\begin{align*}
  {\it describe}(M,g^i,g^o) = \\
  \{ (\rho^i,\rho^o,\pi^i,\pi^o) \mid\ & \rho^i,\pi^i \in {\it parse}(M^i,{\it nil},g^i), \\
                                       & \rho^o,\pi^o \in {\it parse}(M^o,\pi^i,g^o) \}
\end{align*}

The {\em create} mode makes it possible to create a new example of the
task. It consists of the successive generation of an input grid and an
output grid, the latter being conditioned by the former.
This mode is not used in the ARC challenge but it could contribute to
the measurement of the intelligence of a system. Indeed, if an agent
has really understood a task, it should be able to produce new
examples\footnote{At school, teachers often ask pupils to produce
  their own examples of some concept to check their understanding.}.

\begin{align*}
  {\it create}(M) = \{ (\rho^i,\rho^o,g^i,g^o) \mid\ & \rho^i,\pi^i, g^i \in {\it generate}(M^i,{\it nil}), \\
                                       & \rho^o,\pi^o, g^o \in {\it generate}(M^o,\pi^i) \}
\end{align*}

In all modes, the ${\it nil}$ environment is used with the input model
because the input grid comes first, without any prior
information. Note also that all modes inherit the multi-valued
property of parsing and generation.
These three modes highlight an essential difference between our
object-centric models and the DSL-based programs of existing
approaches. The latter are designed for prediction (computation of the
output as a function of the input), they do not provide a description
of the grids, nor a way to create new input grids.
A new example could be created by randomly generating an
input grid and applying the program, but in general, it would not
respect most of the task invariants: e.g., a random bitmap would be
generated rather than a solid square.

\section{MDL-based Model Learning}
\label{learning}

MDL-based learning works by searching for the model that compresses
the data the more. The data to be compressed is here the set of
training examples. We have to define two things: (1) the description
lengths of models and examples, and (2) the search space of models and
the learning strategy.

\subsection{Description Lengths}
\label{dl}

A common approach in MDL is to define the overall description length
(DL) as the sum of two parts ({\em two-parts MDL}): the model~$M$, and
the data~$D$ encoded according to the model~\cite{Grunwald2019}.
\[ L(M,D) = L(M) + L(D\,|\,M) \]
In our case, the model is a task model composed of two grid models,
and the data is the set of training examples (pairs of grids). To
compensate for the small number of examples, and to allow for
sufficiently complex models, we use a {\em rehearsal factor}
$\alpha \geq 1$, like if each example were seen $\alpha$ times.
\begin{align*}
  L(M) =\  & L(M^i) + L(M^o) \\ 
  L(D\,|\,M) =\  & \alpha \sum_{(g^i,g^o)} L(g^i,g^o\,|\,M)
\end{align*}

The DL of an example is based on the most compressive
joint description of the pair of grids. 
\begin{align*}
  L(g^i,g^o\,|\,M) =\ & {\it min}_{\rho^i,\rho^o,\pi^i,\pi^o \in {\it describe}(M,g^i,g^o)} \\
                 & [\  L(\rho^i,\pi^i,g^i\,|\,M^i,{\it nil}) + L(\rho^o,\pi^o,g^o\,|\,M^o,\pi^i) \ ]
\end{align*}
Terms of the form~$L(\rho,\pi,g\,|\,M,\varepsilon)$ denote the DL of a
grid~$g$ encoded according to a grid model~$M$ and an
environment~$\varepsilon$, via the $\rho$-th description~$\pi$
resulting from the parsing, which serves as an intermediate
representation. We can decompose these terms by using $\pi$ as an
intermediate representation of the grid.
\[ L(\rho,\pi,g\,|\,M,\varepsilon) = L(\rho) + L(\pi\,|\,M,\varepsilon) + L(g\,|\,\pi) \]
The term $L(\rho) := L_\nat(\rho) - L_\nat(1)$ encodes the extra-cost
of not choosing the first parsed description, penalizing higher
ranks. $L_\nat(n)$ is a classical {\em universal encoding} for
integers~\cite{elias_encodings}.
The term $L(\pi\,|\,M,\varepsilon)$ measures the amount of information
that must be added to the model and the environment to encode the
description, typically the values of the unknowns.
The term~$L(g\,|\,\pi)$ measures the differences between the original
grid and the grid produced by the description.
A correct model is obtained when $\rho^i = 1$ and
$L(\rho^o,\pi^o,g^o\,|\,M^o,\pi^i) = 0$ for all examples, i.e. when
using the first description for each input grid, there is nothing left
to code for the output grids, and therefore the output grids can be
perfectly predicted from the input grids.

Three elementary model-dependent DLs have to be defined:
\begin{itemize}
\item $L(M)$: DL of a grid model;
\item $L(\pi\,|\,M,\varepsilon)$: DL of a grid description, according
  to the model and environment used for parsing it;
\item $L(g\,|\,\pi)$: DL of a grid, relative to a grid description,
  i.e. the errors commited by the description w.r.t. the grid.
\end{itemize}
We sketch those definitions for the grid models defined in
Section~\ref{models}. We recall that decription lengths are generally
derived from probability distributions with the equation
$L(x) = -\log P(x)$, corresponding to an optimal
coding~\cite{Grunwald2019}.

Defining~$L(M)$ amounts to encode a syntax tree with constructors,
values, unknowns, references, and functions as nodes. Because of
types, only a subset of those are actually possible at each node:
e.g. type {\it Layer} has only one constructor. We use uniform
distributions across possible nodes, and universal encoding for non-bounded ints.
A reference is encoded according to a uniform distribution across all
components of the environment that have a compatible type. We give
unknowns a lower probability than constructors, and
references/functions a higher probability, in order to encourage
models that are more specific, and that make the output depend on the
input.

Defining~$L(\pi\,|\,M,\varepsilon)$ amounts to encode the description
components that are unknowns in the model. As those description
components are actually grounded model components, the above
definitions for~$L(M)$ can be reused, only adjusting the probability
distributions to exclude unknowns, references and functions.

Defining~$L(g\,|\,\pi)$ amounts to encode which cells in grid~$g$ are
wrongly specified by description~$\pi$. For comparability with grid
models and descriptions, we represent each differing cell as a point
object -- {\small {\bf Layer}({\bf Vec}(i,j),{\bf Colored}({\bf
    Point},c))} -- and encode it like descriptions. We also have to
encode the number of differing cells.

\begin{table}[t]
\centering
\caption{Decomposition of $L(M,D)$ for the model in Figure~\ref{fig:model}}
\begin{tabular}{l|rr|r}
  & input & output & pair \\
  \hline
  $L(M)$ & 71.4 & 97.3 & 168.7 \\
  $L(D\,|\,M)$ & 2355.2 & {\bf 0.0} & 2355.2 \\
  \hline
  $L(M,D)$ & 2426.6 & 97.3 & 2523.9 \\
\end{tabular}
\label{tab:dl}
\end{table}

Table~\ref{tab:dl} shows the decomposition of the description
length~$L(M,D)$, for the model in Figure~\ref{fig:model} on
task~\runningtask{}, between the input and the output, and between the
model and the data encoded with the model. Remind that $L(D\,|\,M)$ is
for $\alpha=10$ copies of each example.
It shows that the DL of the output grids is zero bits, which means
that they are entirely determined by the grid models and the input
grids. The proposed model is therefore a solution to the task. The
average DL of an input grid is 2355.2/10/3 = 78.5 bits, on a par with
the DL of the input grid model.

\subsection{Search Space and Strategy}
\label{strategy}

The search space for models is characterized by: (1) an initial model,
and (2) a {\em refinement} operator that returns a list of model
refinements $M_1 \ldots M_n$ given a model~$M$. A refinement can
insert a new component, replace an unknown by a pattern (introducing
new unknowns for the constructor arguments), or replace a model
component by an {\em expression} (a composition of references, values,
and functions). The refinement operator has access to the joint
descriptions, so it can be guided by them.
Similarly to previous MDL-based approaches~\cite{KRIMP2011}, we adopt
a greedy search strategy based on the description length of models. At
each step, starting with the initial model, the refinement that
reduces the more~$L(M,D)$ is selected. The search stops when no model
refinement reduces it.
To compensate for the fact that the input and output grids may have
very different sizes, we actually use a {\em normalized description
  length}~$\hat{L}$ that gives the same weight to the input and output
components of the global DL, relative to the initial model.
\[ \hat{L}(M,D) = \frac{L(M^i,D^i)}{L(M^i_{\it init},D^i)} + \frac{L(M^o,D^o)}{L(M^o_{\it init},D^o)} \in [0,2] \]

Our initial model uses the unknown grid {\tt ?} for both input and
output: $M_{init} = (?,?)$. The available refinements are the following:
\begin{itemize}
\item the insertion of a new layer in the list of layers -- one of
  {\small {\bf Layer}(?,{\bf Col.}({\bf Point},?))}, {\small {\bf
      Layer}(?,{\bf Col.}({\bf Rectangle}(?,?),?))}, {\small {\bf
      Layer}(?,!$object$)}, and {\small !$layer$} -- where !$object$
  (resp. !$layer$) is a reference to an input object (resp. an input
  layer);
\item the replacement of an unknown at path~$p$ by a pattern~$P$
  when for each example, there is a parsed
  description~$\pi$ s.t. $\pi.p$ matches $P$;
\item the replacement of a model component at path~$p$ by an
  expression~$e$ \HIDE{of the same type}when for each example, there is
  a description~$\pi$ s.t. $\pi.p = e$.
\end{itemize}

\begin{table}[t]
  \centering
  \caption{Learning trace for task \runningtask{} ($in.lay[1]$ is inserted
  before $in.lay[0]$ because we use the final layer indices for
  clarity). $in$/$out$ denotes the input/output model, $\hat{L}$ is the normalized DL.}
  {\small
  \setlength\tabcolsep{2pt}
  \begin{tabular}{rlr}
    \hline
    step & refinement & $\hat{L}$ \\
    \hline
    0 & (initial model) & 2.000 \\
    1 & $in \leftarrow {\bf Layers}(?,?,[])$ & 1.117 \\
    2 & $out \leftarrow {\bf Layers}(?,?,[])$ & 0.272 \\
    3 & $in.lay[1] \leftarrow {\bf Layer}(?, {\bf Col.}({\bf Rect.}(?,?),?))$ & 0.179 \\
    4 & $out.lay[0] \leftarrow {\bf Layer}(?, {\bf Col.}({\bf Rect.}(?,?),?))$ & 0.101 \\
    5 & $out.size \leftarrow\ !lay[1].object.shape.size$ & 0.079 \\
    6 & $in.lay[0] \leftarrow {\bf Layer}(?, {\bf Col.}({\bf Rect.}(?,?),?))$ & 0.070 \\
    7 & $out.lay[0].object \leftarrow$ & \\
         & \hspace*{1em} ${\it coloring}(!lay[0].object, !lay[1].object.color)$ & 0.045 \\
    8 & $out.color \leftarrow\ !lay[0].object.color$ & 0.032 \\
    9 & $out.lay[0].pos \leftarrow\ !lay[0].pos - !lay[1].pos$ & 0.020 \\    
    10 & $in.lay[0].object.shape.mask \leftarrow {\bf Full}$ & 0.019 \\
    11 & $in.lay[1].object.shape.mask \leftarrow {\bf Full}$ & 0.019 \\
    12 & $in.color \leftarrow black$ & 0.019 \\
    \hline
  \end{tabular}
}
\label{fig:learning:trace}
\end{table}

Table~\ref{fig:learning:trace} shows the learning trace for task
\runningtask, showing at each step the refinement that was found the
most compressive. It reveals how the system learns about the task
(steps are given in brackets): {\em ``The input and output grids are
  made of layers of objects over a background (1-2). There is a
  rectangle $lay[1]$ in the input (3) and a rectangle $lay[0]$ in the
  output (4). The output grid is the size of $lay[1]$ in input
  (5). There is another rectangle $lay[0]$ in the input, above
  $lay[1]$ (6). We can use its color for the background of the output
  (8). The output rectangle is the same as the input
  rectangle~$lay[0]$ but with the color of $lay[1]$ (7), and its
  position is equal to the difference between the positions of the two
  input rectangles (9). All rectangles are full (10-11) and the input
  background is black (12).''}

\subsection{Pruning Phase}
\label{pruning}

The learned model sometimes lacks generality, and fails on test
examples. This is because the goal of MDL-based learning as defined
above is to find the most compressive task model on pairs of
grids. This is relevant for the description mode, as well as for the
creation mode. However, in the prediction mode, the input grid model
is used as a pattern to match the input grid, and it should be as
general as possible provided that it captures the correct information
for generating the output grid. For example, if all input grids in
training examples have height~10, then the model will fail on a test
example where the input grid has height~12, even if that height does
not matter at all for generating the output.

We therefore add a {\em pruning phase} as a post-processing of the
learned model. The principle is to start from this learned model, and
to repeatdly apply {\em inverse refinements} while this does not break
correct predictions. Inverse refinements can remove a layer or replace
a constructor/value by an unknown. In order to have a uniform learning
strategy, we also use here an MDL-based strategy, only adapting the
description length to the prediction mode.
In prediction mode, the input grid is given, and we therefore replace
$L(g^i,g^o\,|\,M)$ by $L(g^o\,|\,M,g^i)$.  Hence, the DL
$L(\rho^i,\pi^i,g^i\,|\,M^i,nil)$ becomes
$L(\rho^i,\pi^i,g^i\,|\,M^i,nil,g^i)$, which is equal to $L(\rho^i)$
because $\pi^i$ is fully determined by the input grid, the input grid
model, and the parsing rank.
This new definition makes it possible to simplify the input model
while decreasing the DL. Indeed, such simplifications typically reduce
$L(M^i)$ but increase $L(\pi^i|M^i,nil)$ and $L(g^i|\pi^i)$. The two
latter terms are no more counted in the prediction-oriented
measure. The cost related to $\rho^i$ is important because choosing
the wrong description of the input grid almost invariably lead to a
wrong predicted output grid.
The DL $L(\rho^o,\pi^o,g^o\,|\,M^o,\pi^i)$ becomes
$L(\rho^o,\pi^o,g^o\,|\,M^o,\pi^i,g^i)$, which is equal to the former
as $\pi^i$ is an abstract representation of~$g^i$. Indeed, unlike for
input grids, it is important to keep output grids as compressed as
possible.

On task~\runningtask{}, starting from the model in
Figure~\ref{fig:model}, the pruning phase performs three
generalization steps, replacing by unknowns: the {\bf Full} mask of
the two input rectangles, and the {\bf black} color of the input
background. Those generalizations happen not to be necessary on the
test examples of the task in ARC, but they make the model work on
input grids that would break invariants of training examples, e.g. a
cross above a rectangle above a blue background.

\section{Evaluation}
\label{eval}

In this section, we first evaluate our approach on ARC, comparing it
to existing approaches in terms of success rates, efficiency, model
complexity, and model naturalness. We then evaluate the generality of
our approach beyond ARC by applying it to a different domain,
spreadsheets, where inputs and outputs are rows of strings.
Our experiments were run with single-thread implementations on
Fedora~32, Intel Core i7x12 with 16GB memory. We used one run per task
set as there is no randomness involved.

\subsection{Abstraction and Reasoning Corpus (ARC)}
\label{eval:arc}

We evaluated our approach on the 800 public ARC tasks, and we also
took part in the ARCathon~2022 challenge as team MADIL. The few
parameters were set based on the training tasks. To ensure a good
balance of the computational time between parsing and learning, we set
some limits that remained stable across our experiments. The number of
descriptions produced by the parsing of a grid is limited to 64 and
only the 3 most compressive are retained for the computation of
refinements. At each step, at most 100,000 expressions are considered
and only the 20 most promising refinements, according to a DL
estimate, are evaluated. The rehearsal rate~$\alpha$ is set to 10. The
tasks are processed independently of each other, without learning from
one to the other. The results are given for a learning time per task
limited to 60s plus 10s for the pruning phase.

The learning and prediction logs and the screenshots of the solved
training tasks are available as supplementary materials.

\begin{table}[t]
  \centering
  \caption{Number of solved tasks (and percentage) and average
    learning time for solved tasks, for different methods on different
    task sets}
  \begin{tabular}{llrrr}
    task set & method & \multicolumn{2}{l}{solved tasks} & runtime \\
    \hline
    ARC train. & Fischer {\em et al}, 2020 & 31 & 7.68\% & \\
    (400 tasks)         & Alford {\em et al}, 2021 & 22 & 5.50\% & \\
             & Xu {\em et al}, 2022 & 57 & 14.25\% & \\
             & Ainooson {\em et al}, 2023 & 104 & 26.00\% & 178.7s \\
             & OURS & 96 & 24.00\% & 4.6s \\
    \hline
    ARC eval. & Ainooson {\em et al}, 2023 & 26 & 6.50\% & \\
    (400 tasks)         & OURS & 23 & 5.75\% & 11.4s\\
    \hline
    Kaggle'20 & Icecuber (winner) &  & 20.6\% & \\
    (100 tasks)         & Fischer {\em et al} &  & 3.0\% & \\
    \hline
    ARCathon'22 & pablo (winner) & 6 & 6\% & \\
    (100 tasks) & Ainooson {\em et al} & 2 & 2\% & \\
             & OURS (4th ex-aequo) & 2 & 2\% & \\
    \hline
  \end{tabular}
  \label{tab:performance}
\end{table}

{\bf Task sets and baselines.} We consider four task sets for which
results have been reported: the 400 training and 400 evaluation public
tasks, the 100 secret tasks of Kaggle'20, and the 100 secret tasks of
ARCathon'22. We presume that those secret tasks are taken from the 200
secret ARC tasks.
As baselines, we consider published methods that report results on the
considered task
sets~\cite{Fischer2020,Alford2021,Xu2022,Ainooson2023}. We also
include the winners of the two challenges for
reference. Unfortunately, the reported results are scarce, and the
papers do not provide their code.
The code of our method is available as open source on
GitHub\footnote{\url{https://github.com/sebferre/ARC-MDL}}. Version~2.7
was used for the experiments reported here.

{\bf Success rates.} On the training tasks, for which we have the more
results to compare with, our method solves 96 (24\%) training tasks,
almost on par with the best method, by Ainooson {\em et al}
(26\%). Both methods also solved a similar number of evaluation tasks
(23 vs 26 tasks), and both solved 2/100 tasks in ARCathon'22, and
ranked 4th ex-aequo.
Comparing the different task sets, it appears that the evaluation
tasks are significantly more difficult than the training tasks, and
the secret tasks of ARCathon seem even more difficult as the winner
could only solve 6 tasks. Icecuber managed to correctly predict an
amazing 20.6\% of the test output grids in Kaggle'20, but at the cost
of the hand-coding of 142 primitives, 10k lines of code, and
brute-force search (millions of computed grids per task).

The ARC evaluation protocol allows for three predictions per test
example. However, the first prediction of our method is actually
correct in 90 of the 96 solved training tasks. This shows that our
learned models are accurate in their understanding of the tasks.
To better evaluate the generalization capability of learned models, we
also measured the generalization rate as the proportion of models that
are correct on training examples that are also correct on test
examples: 92\% (94/102) on training tasks, and 72\% (23/32) on
evaluation tasks. This again suggests that the evaluation tasks
feature a higher generalization difficulty. Without the pruning phase,
this rate decreases to 89\% (91/102) on training tasks. This shows
that the pruning phase is useful, although description-oriented model
learning is already good at generalization. Reasons for failures to
generalize are: e.g., the test example has several objects while all
training examples have a single object; \HIDE{the parsing of the
  different training examples is inconsistent, which prevents to find
  the correct refinement;} the training examples have a misleading
invariant.



{\bf Efficiency and model complexity.}  Intelligence is the efficiency
at acquiring new skills, according to Chollet. Although ARC enforces
data efficiency by having only a few training examples per task, and
unique tasks, it does not enforce efficiency in the amount of priors,
nor in the computation resources. It is therefore useful to assess the
latter.
We already mentioned Icecuber's method that relies on a large number
of primitives, and intense computations. The method of Ainooson {\em
  et al}, which has comparable performance to ours, uses 52 primitives
and about 700s on average per solved task. In comparison, our method
uses 30 primitives and 4.6s per solved training task (21.7s over all
training tasks). Moreover, doubling the learning timeout at 120s does
not lead to solving more tasks, so 60s just seems to be enough to find
a solution if there is one.
Note also that our method does not stop learning when a solution is
found but when no more compression can be achieved.

Another way to evaluate efficiency is to look at the complexity of
learned models, typically the number of primitives composing the model
in program synthesis approaches. A good proxy for this complexity is
the depth of search that was reached in the allocated time. In our
case, it is equal to the number of refinements applied to the initial
empty model.
Few methods provide this information: Icecuber limits depth to~4, and
Ainooson's best results are achieved with a brute-force search with
maximum depth~3. Methods based on DreamCoder~\cite{Alford2021} have
similar limits but can learn more complex programs by discovering and
defining new operations as common compositions of primitives, and
reusing them from one task to another. Our method can dive much deeper
in less computation time, thanks to its greedy strategy. The number of
refinement steps achieved in a timeout of 60s on the training tasks
ranges from 4 to 57, with an average of 19 steps. This demonstrates
the effectiveness of the MDL criteria to guide the search towards
correct models. This claim is reinforced by the fact that a beam
search (width=3) did not lead to solving more tasks.

{\bf Learned models.}
The learned models for solved tasks are very diverse despite the
simplicity of our models. They express various transformations: e.g.,
moving an object, extending lines, putting one object behind another,
order objects from largest to smallest, remove noise, etc. Note that
none of these transformations is a primitive in our models, they are
learned in terms of objects, basic arithmetics, simple geometry, and
the MDL principle.

We compared our learned models to the natural programs of
LARC~\cite{Acquaviva2022}. Remarkably, many of our models involve
the same objects and similar operations than the natural programs. For
example, the natural program for task \runningtask{} is: {\em ``[The
  input has] a square shape with a small square centered inside the
  large square on a black background. The two squares are of different
  colors. Make an output grid that is the same size as the large
  square. The size and position of the small inner square should be
  the same as in the input grid. The colors of the two squares are
  exchanged.''}
For other tasks, our models miss some notions used by natural programs
but manage to compensate them: e.g., topological relations such as
"next to" or "on top" are compensated by the three attempts; the
majority color is compensated by the MDL principle selecting the
largest object. However, in most cases, the same objects are
identified.

These observations demonstrate that our object-centric models align
well with the natural programs produced by humans, unlike approaches
based on the composition of grid transformations. An example of a
program learned by \cite{Fischer2020} on the task~\task{23b5c85d} is
{\small\tt strip\_black; split\_colors; sort\_Area; top; crop}, which
is a sequence of grid-to-grid transformations, without explicit
mention of objects.

\subsection{From Grids to Strings (FlashFill)}
\label{eval:flashfill}

A similar yet different kind of tasks, compared to ARC, is the
automatic filling of some columns in a spreadsheet given already
filled columns, from only a few input-output examples. A notable work
in program synthesis~\cite{Gulwani2011} has led to a new feature in
Microsoft Excel 2013, called FlashFill. As a simple example, consider
a spreadsheet where column A contains lastnames (e.g., Smith), column
B contains firstnames (e.g., Jones Paul), and column C is expected to
contain the initial of the first firstname followed by the lastname
(e.g., J. Smith).

Like in ARC, each task comes with a few input-output pairs, and the
output should be predicted from the input. The main difference lies in
the type of inputs and outputs, here rows of strings instead of
colored grids. The research hypothesis here is that by changing only
the definition of patterns, functions, and the model-specific DLs, our
approach is able to learn models that solve the tasks given as
examples in the work cited above.

\begin{table}[t]
  \caption{Pattern constructors by type for strings}
  \centering
  \begin{tabular}{|l|l|}
    \hline
    type & constructors \\
    \hline
    \hline
    {\it Row} & {\it Cell}[] \\
    \hline
    {\it Cell} & {\bf Nil} \\
         & {\bf Factor}(left$:${\it Cell}, token$:${\it Token}, right$:${\it Cell}) \\
    \hline
    {\it Token} & {\bf Const}(s$:$string) \\
         & {\bf Regex}(re$:${\it Regex}) \\
    \hline
    {\it Regex} & {\bf Ident}, {\bf Letters}, {\bf Decimal}, {\bf Digits}, {\bf Spaces} \\
    \hline
    $T$ & {\bf Alt}(if$:${\it Cond}, then$:$$T$, else$:$$T$) \\
    \hline 
  \end{tabular}
  \label{fig:string:constructors}
\end{table}

{\bf Models.} Table~\ref{fig:string:constructors} lists the patterns
of our models for rows of strings. A {\em row model} describes a row
of strings, and is simply an array of cell models. A {\em cell model}
decribes the content of a spreadsheet cell, i.e. a string. It is
either the empty string ({\bf Nil}), or the factorization of a string
with a token in the middle and two substrings on each side. Tokens
here play the role of objects. A {\em token model} is either a
constant string or a regular expression, taken among a list of
predefined ones: e.g., {\bf Digits} = {\tt [0-9]+} matches contiguous
sequences of digits. Unknowns (?) can be used as cell models and as
conditions ({\it Cond}). Finally, the {\bf Alt} constructor can be
used in cell models and token models to express alternatives ({\bf
  Alt}(?,$M_1$,$M_2$)), conditionals ({\bf Alt}($expr$,$M_1$,$M_2$)),
and optionals ({\bf Alt}(?,$M$,{\bf Nil})).
The available functions are so far limited: string length, filtering
chars (digits, letters, upper letters, lower letters), converting a
string to uppercase or lowercase, converting ints and bools to
strings, equality to some constant value and logical operators for
conditions. Expressions and references are so far restricted to
tokens.

For comparison, the DSL of FlashFill also uses predefined regular
expressions but uses them to locate positions in the string, rather
than tokens. Their programs are conditional expressions (switch),
where each branch is a concatenation of substrings specified by
position, and constant strings. In contrast, our models allows for
free nestings of conditionals ({\bf Alt}) and concatenation ({\bf
  Factor}). However, their DSL has loops that have so far no
counterpart in our models.

{\bf Task set.}
For a preliminary evaluation, we used as a task set the 14 examples
in~\cite{Gulwani2011}. Each task has one or two strings as inputs and
one string as output, and 2-6 training examples (avg. 3.4). We
complement them with 3-6 evaluation examples, some of which feature
some generalization difficulty. Those 14 tasks are available in the
supplementary materials in the same JSON format as ARC tasks.

{\bf Efficiency and success rates.}
Learning takes 1s or less, except for Task~1 and Task~13 that have
longer input strings and take respectively 9.9s and 5.2s. The depth of
search ranges from 11 to 76, and averages at 35 steps.
For 11/14 tasks (all except Tasks 4, 5, 9), the learned model
correctly describes and predicts the training examples. However, only
5 of those learned models generalize to all test examples: 3 models
fail on a single test example (e.g., in Task~3 the file extension {\tt
  .mp4} contains a digit unlike other file extensions); in Task~8, the
training examples are ambiguous because the input string is a date in
different formats, and the output string could either be the day or
last two digits of the year; in Task~11, there is a typo in the
training examples (on purpose), which makes the task under-specified.
The (partial) failure for other tasks is explained by missing features
in our models, notably the counterpart of loops, or by a wrong
sequence of refinements. For instance, Task~9 can be solved by
delaying the insertion of alternatives.

{\bf Learned models.}  Input models can be expressed as regular
expressions with groups on tokens and alternatives, and output models
can be expressed as string interpolations with group identifiers as
variables. For Task~10, we obtain the following model.

\noindent
$M^i$: {\small \verb#\(.*\([0-9]+\).*\)?\([0-9]+\).*\([0-9]+\)#} \\
$M^o$: {\small \verb#{if \1 then \2 else "425"}-\3-\4#}

The input is made of three integers, the first one being optional. The
output is the concatenation of those three integers, separated by
dashes, and the first integer is 425 when missing in the input.
In FlashFill, in general, a large number of programs is generated as
an exhaustive search is performed. For Task~10, the program given as
solution in~\cite{Gulwani2011} is the following ($A$ refers to the
input column):

{\footnotesize
\[\begin{array}{l}
  Switch((b1,e1),(b2,e2)), where\\
    b1 \equiv Match(A,NumTok,3), \\
    b2 \equiv \neg Match(A,NumTok,3), \\
    e1 \equiv Concatenate(SubStr2(A,NumTok,1),\\
    \hspace{1cm} Const(\textrm{"-"}), SubStr2(A,NumTok,2), \\
    \hspace{1cm} Const(\textrm{"-"}), SubStr2(A,NumTok,3))\\
    e2 \equiv Concatenate(Const(\textrm{"425-"}), SubStr2(A,NumTok,1),\\
    \hspace{1cm} Const(\textrm{"-"}), SubStr2(A,NumTok,2))
  \end{array}\]}
This is illustrative of the different programming style between our
pattern-based models and the computation-based DSL programs.

\section{Conclusion and Perspectives}
\label{conclu}

We have presented a novel and general approach to efficiently learn
skills at tasks that consist in generating structured outputs as a
function of structured inputs. Following Chollet's measure of
intelligence, {\em efficiently learning} here means limited knowledge
prior for the target scope of tasks, only a few examples per task, and
low computational resources. Our approach is based on descriptive task
models that combine object-centric patterns and computations, and on
the MDL principle for guiding the search for models. We have detailed
an application to ARC tasks on colored grids, and sketched an
application to FlashFill tasks on strings. We have shown promising
results, especially in terms of efficiency, model complexity, and
model naturalness.

Going further on ARC will require a substantial design effort as our
current models cover so far a small subset of the knowledge priors
that are required by ARC tasks (e.g., goal-directedness). For
FlashFill tasks, the addition of a counterpart for loops and common
functions is expected to suffice to match the state-of-the-art.



\section*{Supplementary Materials}

We here describe the contents of the supplementary file accompanying
the paper. A public repository of the source code is also available at
\url{https://github.com/sebferre/ARC-MDL} for ARC and at
\url{https://github.com/sebferre/ARC-MDL-strings} for FlashFill.
There are two main directories: one for ARC tasks and another for
FlashFill tasks.

\subsection*{Task Sets}

The public task sets of ARC are available online at
\url{https://github.com/fchollet/ARC}. There are two task sets:
training tasks and evaluation tasks, each containing 400 tasks.

The task set of FlashFill is made of the 14 examples
in~\cite{Gulwani2011}. We provide them as JSON files in {\tt
  FlashFill/taskset/}, using the same format as ARC tasks, except that
strings and arrays of strings are used instead of colored grids. For
convenience, we also provide the file {\tt
  FlashFill/taskset/all\_examples.json} to allow for browsing all
examples in one file.

\subsection*{Results}

We provide the learning and prediction logs for each task set:
\begin{itemize}
\item {\tt ARC/training\_tasks.log}
\item {\tt ARC/evaluation\_tasks.log}
\item {\tt FlashFill/tasks.log}
\end{itemize}
Each log file starts with the hyperparameter values, and ends with
global statistical measures. For each task, it gives:
\begin{itemize}
\item the detailed DL (description length) of the initial model;
\item the learning trace (including the pruning phase) as a sequence
  of refinements, and showing the decrease of the normalized DL;
\item the learned models before and after pruning and their detailed
  DL;
\item the best joint description for each training example, except for
  ARC evaluation tasks so as not to leak their contents to the AI
  developer (a recommendation made by F. Chollet);
\item the prediction for each training and test example;
\item and finally a few measures for the task.
\end{itemize}
  
The measures given for each task and at the end are the
following:
\begin{itemize}
\item {\tt runtime-learning}: learning time in seconds (including the pruning phase);
\item {\tt bits-train-error}: the remaining error commited on output training grids, in bits;
\item {\tt acc-train-micro}: the proportion of training output grids that are correctly predicted;
\item {\tt acc-train-macro}: 1 if all training output grids are correctly predicted, 0 otherwise;
\item {\tt acc-train-mrr}: Mean Reciprocal Rank (MRR) of correct predictions for training output grids, 1 if all first predictions are correct;
\item {\tt acc-test-micro}: the proportion of test output grids that are correctly predicted;
\item {\tt acc-test-macro}: 1 if all test output grids are correctly predicted, 0 otherwise;
\item {\tt acc-test-mrr}: Mean Reciprocal Rank (MRR) of correct predictions for test output grids, 1 if all first predictions are correct.
\end{itemize}
The reference measure in ARC is {\tt acc-test-macro}. The micro
measures provide a more fine-grained and more optimistic measure of
success.

For convenience, we also provide in {\tt ARC/solved\_tasks} a picture
for each of the 96 training ARC tasks that are solved by our
approach. We kindly invite the reader to browse them to get a quick
idea of the diversity of the tasks that our approach can solve. The
pictures are screenshots from the UI provided along with ARC tasks.

\end{document}